\documentclass[12pt,a4paper]{amsart}
\usepackage{amsfonts}
\usepackage{amsthm}
\usepackage{amsmath}
\usepackage{amscd}
\usepackage[latin2]{inputenc}
\usepackage{t1enc}
\usepackage[mathscr]{eucal}
\usepackage{indentfirst}
\usepackage{graphicx}
\usepackage{graphics}
\usepackage{pict2e}
\usepackage{epic}
\numberwithin{equation}{section}
\usepackage[margin=2.9cm]{geometry}
\usepackage{epstopdf}

\theoremstyle{plain}
\newtheorem{Th}{Theorem}[section]
\newtheorem{Lemma}[Th]{Lemma}

\newtheorem{Prop}[Th]{Proposition}
\newtheorem{Assump}[Th]{Assumption}

 \theoremstyle{definition}

\newtheorem{?}[Th]{Problem}

\begin{document}

\title[Convergence for stochastic shortest path]{On the Convergence of Optimistic Policy Iteration for the Stochastic Shortest Path Problem}

\author[Y. Chen]{Yuanlong Chen}

\address{University of Washington, Seattle}

\email{ylchen88@uw.edu}


\begin{abstract}
In this paper, we prove convergence results for a special case of the optimistic policy iteration algorithm for the stochastic shortest path problem considered in \cite{Ts03}. We consider both Monte Carlo and $TD(\lambda)$ methods for the policy evaluation step under the assumption that the terminal state is reached almost surely under every policy.
\end{abstract}

\maketitle

\section{Introduction}
In this paper, we consider a Markov decision process (MDP) with the finite state set $S=\{1,2,\dots,n\}$. In addition, we use $0$ to denote a cost-free absorbing terminal state. For each state $i$, we assume that there are finitely many actions, denoted by $U(i)$. Furthermore, for each state $i\in S$, action $u\in U(i)$, and state $j\in S\cup\{0\}$, we associate a transition probability $p_{i,j}(u)$ and an immediate cost $g(i,u)$. A policy $\mu$ is a mapping satisfying $\mu(i)\in U(i)$ for every $i\in S$. Note that there are only finitely many policies because the state and action sets are finite. Let $X_t^{\mu}$ denote the state at time $t$ under policy $\mu$. Then $\{X_t^{\mu}\}$ forms a Markov chain with transition probabilities
$$
P(X_{t+1}^{\mu}=j\mid X_t^{\mu}=i)=p_{i,j}(\mu(i)).
$$
The total expected cost (cost-to-go) of the process starting from state $i$ under policy $\mu$ is
$$
J^{\mu}(i)=E\left[\sum_{t=0}^{\infty}\alpha^t g(X_t^{\mu},\mu(X_t^{\mu}))\,\big|\,X_0^{\mu}=i\right],
$$
where $0<\alpha\leq 1$ is the discount factor. A policy $\mu$ is said to be proper if, under this policy, the terminal state has positive probability of being reached within at most $n$ steps, regardless of the initial state; equivalently, if
$$
\rho_{\mu}=\max_{i\in S}P(X_n^{\mu}\neq 0\mid X_0^{\mu}=i)<1.
$$
A proper policy implies that the terminal state is eventually reached almost surely. To see this, note that
$$
P(X_t^{\mu}\neq 0\mid X_0^{\mu}=i)\leq \rho_{\mu}^{\lfloor t/n\rfloor},\qquad \forall i\in S.
$$
The conclusion follows directly by letting $t\to\infty$ in the preceding bound. Moreover, $J^{\mu}$ is finite when $\mu$ is proper, because
$$
|J^{\mu}(i)|\leq \lim_{T\to\infty}\sum_{t=0}^{T-1}\rho_{\mu}^{\lfloor t/n\rfloor}\max_j|g(j,\mu(j))|<\infty,\qquad \forall i\in S.
$$

In this paper, we assume that every policy is proper.
\begin{Assump} \label{assumption}
	Every policy in our problem is proper. 
\end{Assump}

Let
$$
\bar g=\max_{i\in S,\,u\in U(i)}|g(i,u)|
\qquad\text{and}\qquad
\rho=\max_{\mu}\rho_{\mu}.
$$
Since the policy set is finite, $\rho<1$. Lemma~\ref{lmtau} gives a uniform second-moment bound for the termination times.

In the remainder of the paper, we consider only the stochastic shortest path case $\alpha=1$. We denote the optimal cost-to-go function starting from state $i$ by $J^{*}(i)$, namely, the minimum of the cost-to-go functions over all policies:
$$
J^{*}(i)=\min_{\mu}J^{\mu}(i).
$$
For each state, the minimum is attained because there are only finitely many policies. We define the optimal cost-to-go vector by $J^{*}=(J^{*}(1),\dots,J^{*}(n))$. A policy $\mu$ is said to be optimal if $J^{\mu}(i)=J^{*}(i)$ for every $i\in S$.

We next introduce two dynamic programming operators. For any $n$-dimensional vector $J=(J(1),\dots,J(n))$, define the operator $T:\mathbb{R}^n\to\mathbb{R}^n$ by
$$
(TJ)(i)=\min_{u\in U(i)}\left\{g(i,u)+\sum_{j=1}^{n}p_{i,j}(u)J(j)\right\},\qquad \forall i\in S.
$$
Similarly, define $T_{\mu}:\mathbb{R}^n\to\mathbb{R}^n$ by
$$
(T_{\mu}J)(i)=g(i,\mu(i))+\sum_{j=1}^{n}p_{i,j}(\mu(i))J(j),\qquad \forall i\in S.
$$
In vector notation, these definitions are equivalent to
$$
(TJ)(i)=\min_{\mu}(T_{\mu}J)(i),\qquad \forall i\in S,
$$
and
$$
T_{\mu}J=g_{\mu}+P_{\mu}J.
$$
These two operators have several well-known properties, which we summarize in the following proposition; see \cite{Be00,BeT89,BeT91}.
\begin{Prop}\label{propdp}
Under Assumption~\ref{assumption}, the following properties hold for the stochastic shortest path problem:
\begin{enumerate}
\item[(a)] The optimal cost-to-go vector $J^{*}$ has finite components and satisfies
$$
J^{*}=TJ^{*}.
$$
Furthermore, $J^{*}$ is the unique solution of this equation.
\item[(b)] For every vector $J$,
$$
\lim_{k\to\infty}T^kJ=J^{*}.
$$
\item[(c)] A policy $\mu$ is optimal if and only if
$$
T_{\mu}J^{*}=TJ^{*}.
$$
\item[(d)] For every proper policy $\mu$ and every vector $J$,
$$
\lim_{k\to\infty}T_{\mu}^{k}J=J^{\mu}.
$$
Furthermore,
$$
J^{\mu}=T_{\mu}J^{\mu},
$$
and $J^{\mu}$ is the unique solution of this equation.
\end{enumerate}
\end{Prop}

Throughout this paper, for an $n$-dimensional vector $J$, we use $\|\cdot\|$ to denote the maximum norm, defined by
$$
\|J\|=\max_i|J(i)|.
$$
For an $n$-dimensional vector $\xi=(\xi(1),\dots,\xi(n))$ with positive components, we use $\|\cdot\|_{\xi}$ to denote the weighted maximum norm with respect to $\xi$, defined by
$$
\|J\|_{\xi}=\max_i\frac{|J(i)|}{\xi(i)}.
$$
For two vectors $J$ and $\bar J$, we write $J\leq\bar J$ if $J(i)\leq\bar J(i)$ for every $i\in S$. Vector inequalities throughout the paper are interpreted componentwise.

Under Assumption~\ref{assumption}, Proposition 2.2 on p.~23 of \cite{BertsekasTsitsiklis96} implies that there exist a vector $\xi\in\mathbb{R}^n$ with positive components and a scalar $\beta\in[0,1)$ such that
\begin{equation}\label{weightedcontraction}
\sum_{j=1}^{n}p_{i,j}(u)\xi(j)\leq\beta\xi(i),\qquad \forall i\in S,\quad \forall u\in U(i).
\end{equation}
Consequently, for every policy $\mu$ and all vectors $J,\bar J$,
$$
\|T_{\mu}J-T_{\mu}\bar J\|_{\xi}\leq\beta\|J-\bar J\|_{\xi},
\qquad
\|TJ-T\bar J\|_{\xi}\leq\beta\|J-\bar J\|_{\xi}.
$$

We also use the following monotonicity properties of $T$ and $T_{\mu}$; see Lemma 2.1 in \cite{BertsekasTsitsiklis96}.
\begin{Prop}\label{mon}
For all $n$-dimensional vectors $J$ and $\bar J$ satisfying $J\leq\bar J$, every policy $\mu$, and every positive integer $k$,
$$
T^kJ\leq T^k\bar J,\qquad T_{\mu}^kJ\leq T_{\mu}^k\bar J.
$$
\end{Prop}
Let $e$ denote the $n$-dimensional vector whose components are all equal to $1$. The following result is a direct consequence of induction and Proposition~\ref{mon}.
\begin{Lemma}\label{lmmon}
For every nonnegative scalar $c$, every vector $J$, every policy $\mu$, and every positive integer $k$,
$$
T^k(J+ce)\leq T^kJ+ce,
$$
$$
T_{\mu}^k(J+ce)\leq T_{\mu}^kJ+ce.
$$
If $c<0$, the two displayed inequalities hold with the inequality signs reversed.
\end{Lemma}
\begin{proof}
For $k=1$, the result follows from
$$
T_{\mu}(J+ce)=T_{\mu}J+cP_{\mu}e
$$
and $P_{\mu}e\leq e$. The same componentwise argument, applied before taking the minimum over actions, gives the corresponding inequality for $T$. When $c<0$, multiplication by $c$ reverses the inequalities. The result for every $k\geq1$ then follows by induction using Proposition~\ref{mon}.
\end{proof}
For $T_{\mu}$, we also have the following lemma.
\begin{Lemma}\label{lmtmu}
Let $\{\lambda_l\}_{l=0}^{\infty}$ be a scalar sequence satisfying $\lambda_l\geq 0$ and $\sum_l\lambda_l=1$. For any bounded vector sequence $\{J_l\}_{l=0}^{\infty}$,
$$
T_{\mu}\left(\sum_{l=0}^{\infty}\lambda_lJ_l\right)=\sum_{l=0}^{\infty}\lambda_lT_{\mu}J_l.
$$
\end{Lemma}
\begin{proof}
Because $\{J_l\}$ is bounded and $\sum_l\lambda_l=1$, the vector series is absolutely convergent. Using $T_{\mu}J=g_{\mu}+P_{\mu}J$, we obtain
$$
\begin{aligned}
T_{\mu}\left(\sum_{l=0}^{\infty}\lambda_lJ_l\right)
&=g_{\mu}+P_{\mu}\sum_{l=0}^{\infty}\lambda_lJ_l\\
&=\sum_{l=0}^{\infty}\lambda_l\bigl(g_{\mu}+P_{\mu}J_l\bigr)
=\sum_{l=0}^{\infty}\lambda_lT_{\mu}J_l.
\end{aligned}
$$
\end{proof}

We now give a brief description of the policy iteration algorithm. In ordinary policy iteration, we start with an initial policy $\mu$ and perform policy evaluation, that is, we evaluate the cost-to-go vector $J^{\mu}$ corresponding to $\mu$. For example, one can use learning algorithms such as Monte Carlo or $TD(\lambda)$ in this step. Once $J^{\mu}$ is available, we perform a policy improvement step, which updates $\mu$ according to
$$
\mu(i)\leftarrow\arg\min_{u\in U(i)}\left\{g(i,u)+\sum_{j=1}^{n}p_{i,j}(u)J^{\mu}(j)\right\},\qquad \forall i\in S.
$$
This process is repeated until the algorithm converges.

One disadvantage of the algorithm described above is that, in practice, accurately evaluating $J^{\mu}$ can be expensive and can make the algorithm inefficient. Optimistic policy iteration is a variation of ordinary policy iteration that addresses this issue by basing policy improvement on an incomplete evaluation of $J^{\mu}$. For example, if Monte Carlo evaluation is used, ordinary policy iteration theoretically requires a large number of simulated trajectories to obtain an accurate estimate. In contrast, optimistic policy iteration performs policy improvement immediately after a single trajectory sample. In \cite{Ts03}, convergence was established for discounted problems ($0<\alpha<1$) using both Monte Carlo and $TD(\lambda)$ policy evaluation. In the following sections, we show that similar convergence results extend to the undiscounted stochastic shortest path problem ($\alpha=1$).

\section{Monte Carlo-based optimistic policy iteration}
We first provide a precise description of the optimistic policy iteration algorithm. We start with an arbitrary, possibly random, vector $J_0$. At each iteration $t$, let $\mu_t$ be a greedy policy with respect to $J_t$, selected using a fixed tie-breaking rule, so that
$$
T_{\mu_t}J_t=TJ_t.
$$
For each state $i$, we simulate a single trajectory starting from $i$ under policy $\mu_t$; termination is guaranteed because $\mu_t$ is proper. Let $R_t(i)$ denote the observed cumulative cost and write
$$
R_t(i)=J^{\mu_t}(i)+\omega_t(i).
$$
We then update $J_t$ according to
\begin{equation}\label{updaterule}
J_{t+1}(i)=(1-\gamma_t)J_t(i)+\gamma_t\bigl(J^{\mu_t}(i)+\omega_t(i)\bigr),
\end{equation}
where $\gamma_t$ is a deterministic scalar stepsize. We assume
$$
0<\gamma_t\leq 1,\qquad
\sum_{t=0}^{\infty}\gamma_t=\infty,\qquad
\sum_{t=0}^{\infty}\gamma_t^2<\infty.
$$

Let $\mathcal{F}_t$ be the history of the algorithm up to and including the point at which $J_t$ has been produced, but before the trajectories for the next update are simulated. Conditional on $\mathcal{F}_t$, the trajectories at iteration $t$ are generated under $\mu_t$ using fresh simulation randomness.

Lemma~\ref{lmmcnoise} shows that the Monte Carlo noise is a martingale-difference sequence with a uniform conditional second-moment bound.

Lemma~\ref{lmcompare} provides the localization and comparison principle used below.

We summarize the main result in the following theorem.

\begin{Th}\label{main}
The sequence $J_t$ generated by the optimistic policy iteration algorithm according to \eqref{updaterule} for the stochastic shortest path problem converges almost surely to the optimal cost-to-go vector $J^*$.
\end{Th}

Before proving Theorem~\ref{main}, we establish several preliminary results.

\begin{Lemma}\label{lmapprox}
For any $\epsilon>0$ and $M>0$, there exists a positive integer $K=K(\epsilon,M)$ such that, for every policy $\mu$, every vector $J$ satisfying $\|J\|\leq M$, and every $k\geq K$,
$$
\|T_{\mu}^kJ-J^{\mu}\|<\epsilon.
$$
\end{Lemma}
\begin{proof}
Fix a policy $\mu$. For every $n$-dimensional vector $J$, part (d) of Proposition~\ref{propdp} gives
$$
\lim_{k\to\infty}T_{\mu}^kJ=J^{\mu}.
$$
Thus, for every $\epsilon>0$, there exists $K(J)>0$ such that
$$
\|T_{\mu}^kJ-J^{\mu}\|<\epsilon/2,\qquad \forall k\geq K(J).
$$
Moreover,
$$
\|T_{\mu}J-T_{\mu}\bar J\|\leq\|J-\bar J\|,
$$
and induction gives
$$
\|T_{\mu}^kJ-T_{\mu}^k\bar J\|\leq\|J-\bar J\|,\qquad \forall k\geq 1.
$$
Therefore, if $\|\bar J-J\|<\epsilon/2$, then
$$
\|T_{\mu}^k\bar J-J^{\mu}\|<\epsilon,\qquad \forall k\geq K(J).
$$
Let $R=\{J:\|J\|\leq M\}$. The sets $B_{\epsilon}(J)=\{\bar J:\|\bar J-J\|<\epsilon/2\}$ form an open cover of the compact set $R$. By the Heine--Borel theorem, there is a finite subcover $B_{\epsilon}(J_1),\dots,B_{\epsilon}(J_l)$. Setting
$$
K_{\mu}=\max_{1\leq r\leq l}K(J_r)
$$
gives the desired bound for this fixed policy. Finally, because there are only finitely many policies, taking the maximum of $K_{\mu}$ over all policies gives a common $K(\epsilon,M)$.
\end{proof}

\begin{Lemma}\label{lmbounded}
The sequence $J_t$ generated by the optimistic policy iteration algorithm according to \eqref{updaterule} is bounded almost surely.
\end{Lemma}
\begin{proof}
Because there are only finitely many policies, the vectors $J^{\mu_t}$ are uniformly bounded. The update rule is
$$
J_{t+1}=(1-\gamma_t)J_t+\gamma_tJ^{\mu_t}+\gamma_t\omega_t.
$$
Lemma~\ref{lmmcnoise} supplies the required uniform conditional second-moment bound, and the almost-sure boundedness of $\{J_t\}$ follows from Proposition 4.7 on p.~159 of \cite{BertsekasTsitsiklis96}.
\end{proof}

Define the scalar sequences $c_t$ and $c_t^+$ by
\begin{equation}\label{dfct}
c_t=\max_i\bigl((TJ_t)(i)-J_t(i)\bigr),
\qquad
c_t^+=\max\{c_t,0\}.
\end{equation}
\begin{Lemma}\label{lmct}
The sequence $c_t$ satisfies
$$
\limsup_{t\to\infty}c_t\leq 0,
$$
and consequently $c_t^+\to 0$.
\end{Lemma}
\begin{proof}
Let $r_t=TJ_t-J_t$. Recall that
$$
T_{\mu_t}J=g_{\mu_t}+P_{\mu_t}J,
\qquad \forall J.
$$
Using $T_{\mu_t}J_t=TJ_t$, $T_{\mu_t}J^{\mu_t}=J^{\mu_t}$, and the affine form of $T_{\mu_t}$, we obtain
$$
r_{t+1}\leq(1-\gamma_t)r_t+\gamma_tv_t,
$$
where $v_t=P_{\mu_t}\omega_t-\omega_t$. By Lemma~\ref{lmmcnoise},
$$
E[v_t\mid\mathcal{F}_t]=0.
$$
Moreover, \eqref{weightedcontraction} gives
$$
\|v_t\|_{\xi}\leq(1+\beta)\|\omega_t\|_{\xi},
$$
so equivalence of norms and Lemma~\ref{lmmcnoise} imply
$$
E[\|v_t\|_{\xi}^2\mid\mathcal{F}_t]\leq C'
$$
for some constant $C'>0$. Lemma~\ref{lmbounded} implies that $\{r_t\}$ is almost surely bounded. Apply Lemma~\ref{lmcompare} with
$$
X_t=r_t,\qquad Q_t=0,\qquad \Phi\equiv0,\qquad \sigma=0.
$$
We obtain
$$
\limsup_{t\to\infty}r_t\leq0
$$
componentwise. Hence $\limsup_{t\to\infty}c_t\leq0$, and the definition of $c_t^+$ implies $c_t^+\to0$.
\end{proof}

\begin{Lemma}\label{lmjmuest}
Almost surely, for every $\epsilon>0$, there exists a finite time $t(\epsilon)$ such that, for all $t\geq t(\epsilon)$,
\begin{equation}\label{keyest}
J^{\mu_t}\leq TJ_t+\epsilon e.
\end{equation}
\end{Lemma}
\begin{proof}
Fix $\epsilon>0$. For each positive integer $m$, consider the event
$$
A_m=\left\{\sup_t\|J_t\|\leq m\right\}.
$$
By Lemma~\ref{lmbounded}, $\bigcup_{m\geq1}A_m$ has probability one. On $A_m$, choose
$$
K=K(\epsilon/2,m)
$$
as in Lemma~\ref{lmapprox}. Since $\mu_t$ is greedy,
$$
T_{\mu_t}J_t=TJ_t\leq J_t+c_t^+e.
$$
Lemma~\ref{lmmon} and induction therefore give
\begin{equation}\label{finitehorizonresidual}
T_{\mu_t}^{k+1}J_t
\leq TJ_t+k c_t^+e,
\qquad \forall k\geq0.
\end{equation}
Lemma~\ref{lmapprox} gives
$$
J^{\mu_t}\leq T_{\mu_t}^{K}J_t+\frac{\epsilon}{2}e.
$$
Applying $T_{\mu_t}$ to both sides and using Lemma~\ref{lmmon}, followed by \eqref{finitehorizonresidual}, yields
$$
J^{\mu_t}
\leq T_{\mu_t}^{K+1}J_t+\frac{\epsilon}{2}e
\leq TJ_t+\left(Kc_t^++\frac{\epsilon}{2}\right)e.
$$
By Lemma~\ref{lmct}, $c_t^+\to0$ almost surely. Thus, on $A_m$, there is a finite random time after which $Kc_t^+\leq\epsilon/2$, and hence
$$
J^{\mu_t}\leq TJ_t+\epsilon e.
$$
Taking the countable union over $m$ proves the assertion for the fixed $\epsilon$. Taking a countable intersection over positive rational $\epsilon$ gives a single probability-one event on which the assertion holds for every $\epsilon>0$.
\end{proof}

\begin{proof}[Proof of Theorem~\ref{main}]
Fix $\epsilon>0$. By Lemma~\ref{lmjmuest}, for all sufficiently large $t$,
$$
\begin{aligned}
J_{t+1}
&=(1-\gamma_t)J_t+\gamma_tJ^{\mu_t}+\gamma_t\omega_t\\
&\leq(1-\gamma_t)J_t+\gamma_tTJ_t+\gamma_t\epsilon e+\gamma_t\omega_t.
\end{aligned}
$$
Define $H_{\epsilon}:\mathbb{R}^n\to\mathbb{R}^n$ by
$$
H_{\epsilon}J=TJ+\epsilon e.
$$
By \eqref{weightedcontraction}, $H_{\epsilon}$ is a contraction in $\|\cdot\|_{\xi}$ with modulus $\beta$. Let $J_{\epsilon}$ be its unique fixed point. Since $J^*=TJ^*$,
$$
\begin{aligned}
\|J_{\epsilon}-J^*\|_{\xi}
&=\|TJ_{\epsilon}+\epsilon e-TJ^*\|_{\xi}\\
&\leq\beta\|J_{\epsilon}-J^*\|_{\xi}+\epsilon\|e\|_{\xi},
\end{aligned}
$$
so
\begin{equation}\label{perturbedfixedpoint}
\|J_{\epsilon}-J^*\|_{\xi}
\leq\frac{\epsilon\|e\|_{\xi}}{1-\beta}.
\end{equation}

Lemma~\ref{lmbounded}, Lemma~\ref{lmmcnoise}, and Lemma~\ref{lmcompare}, applied with
$$
X_t=J_t,\qquad Q_t=0,\qquad \Phi=H_{\epsilon},\qquad \sigma=t(\epsilon),
$$
give
$$
\limsup_{t\to\infty}J_t\leq J_{\epsilon}.
$$
Applying the preceding comparison for $\epsilon=1/m$, $m\geq1$, and taking the countable intersection of the resulting probability-one events, we may let $m\to\infty$ in \eqref{perturbedfixedpoint}. This yields
$$
\limsup_{t\to\infty}J_t\leq J^*.
$$

For the reverse inequality, $J^{\mu_t}\geq J^*$ for every policy $\mu_t$. Hence
$$
J_{t+1}\geq(1-\gamma_t)J_t+\gamma_tJ^*+\gamma_t\omega_t.
$$
Apply the lower-comparison part of Lemma~\ref{lmcompare} with
$$
X_t=J_t,\qquad Q_t=0,\qquad \Phi(J)=J^*,\qquad \sigma=0.
$$
Since the constant mapping $\Phi$ is a monotone contraction with fixed point $J^*$, we obtain
$$
\liminf_{t\to\infty}J_t\geq J^*.
$$
Combining the upper and lower bounds proves that $J_t\to J^*$ almost surely.
\end{proof}

\section{$TD(\lambda)$-based optimistic synchronous policy iteration}
In this section, we extend the result of the previous section to optimistic policy iteration with $TD(\lambda)$ policy evaluation. The algorithm is the same as the Monte Carlo algorithm described above, except that the policy evaluation step uses temporal differences. At iteration $t$, let $J_t$ be the current vector and let $\mu_t$ be a corresponding greedy policy, selected using the same fixed tie-breaking rule. For each state $i$, simulate a trajectory $i_0,i_1,\dots$ with $i_0=i$, and update $J_t(i)$ according to
$$
J_{t+1}(i)=J_t(i)+\gamma_t\sum_{k=0}^{\infty}\lambda^kd_k,
\qquad \lambda\in[0,1),
$$
where
$$
d_k=g(i_k,\mu_t(i_k))+J_t(i_{k+1})-J_t(i_k)
$$
is the temporal difference and $\gamma_t$ is a scalar stepsize. We take $J_t(0)=0$ and all costs after termination to be zero. The update is equivalent to
$$
\begin{aligned}
J_{t+1}(i)
=(1-\gamma_t)J_t(i)
+\gamma_t(1-\lambda)\sum_{k=0}^{\infty}\lambda^k
\bigg(&\sum_{l=0}^{k}g(i_l,\mu_t(i_l))\\
&+J_t(i_{k+1})\bigg).
\end{aligned}
$$
Let $\widehat F_{\lambda,t}(i)$ denote the sampled target in the preceding display, and let $\widehat F_{\lambda,t}$ be the corresponding vector. In vector notation, write
$$
\widehat F_{\lambda,t}=F_{\lambda}^{\mu_t}J_t+\omega_t,
$$
where
\begin{equation}\label{Flambda}
F_{\lambda}^{\mu}J=(1-\lambda)\sum_{k=0}^{\infty}\lambda^kT_{\mu}^{k+1}J.
\end{equation}
The series is well defined because Proposition~\ref{propdp}(d) implies that $\{T_{\mu}^{k+1}J\}$ is bounded. Then the update becomes
\begin{equation}\label{updaterule2}
J_{t+1}=(1-\gamma_t)J_t+\gamma_tF_{\lambda}^{\mu_t}J_t+\gamma_t\omega_t.
\end{equation}
We use the same stepsizes and fresh trajectory simulations as in Section~2.
Lemma~\ref{lmtdnoise} establishes the conditional-expectation identity defining $F_{\lambda}^{\mu_t}J_t$ and the conditional second-moment bound for the sampling noise.

Before stating the main result, consider the two endpoint cases. Although \eqref{updaterule2} is written for $\lambda<1$, setting $\lambda=1$ in the temporal-difference sum above makes the sum telescope and gives
$$
J_{t+1}(i)=(1-\gamma_t)J_t(i)+\gamma_t\sum_{k=0}^{\infty}g(i_k,\mu_t(i_k)),
$$
which is the Monte Carlo method. At the other endpoint, if $\lambda=0$, then
$$
J_{t+1}=(1-\gamma_t)J_t+\gamma_tTJ_t+\gamma_t\omega_t,
$$
because $T_{\mu_t}J_t=TJ_t$. Under Assumption~\ref{assumption}, $T$ is a weighted maximum-norm contraction by \eqref{weightedcontraction}; Proposition 4.4 on p.~156 of \cite{BertsekasTsitsiklis96} therefore implies convergence to $J^*$. For $0<\lambda<1$, the method combines features of $TD(0)$ and Monte Carlo evaluation. We show below that it also converges almost surely to $J^*$.

\begin{Th}\label{main2}
For every fixed $\lambda\in[0,1]$, the sequence $J_t$ generated by the $TD(\lambda)$-based optimistic synchronous policy iteration algorithm converges almost surely to the optimal cost-to-go vector $J^*$. For $\lambda<1$, the trajectory-sampling update is given by \eqref{updaterule2}; at $\lambda=1$, the temporal-difference sum telescopes to the Monte Carlo update of Section~2.
\end{Th}

The case $\lambda=0$ was handled above. In the lemmas below, we therefore assume $0<\lambda<1$ and establish results parallel to Lemmas \ref{lmbounded}--\ref{lmjmuest}.

\begin{Lemma}\label{lmbounded2}
The sequence $J_t$ generated by the optimistic policy iteration algorithm according to \eqref{updaterule2} is bounded almost surely.
\end{Lemma}
\begin{proof}
Let
$$
q_{\lambda}=(1-\lambda)\sum_{k=0}^{\infty}\lambda^k\beta^{k+1}
=\frac{(1-\lambda)\beta}{1-\lambda\beta}.
$$
Because $0<\lambda<1$ and $0\leq\beta<1$, we have $0\leq q_{\lambda}<1$. For every policy $\mu$, Proposition~\ref{propdp}(d) and the affine form of $T_{\mu}$ give
$$
T_{\mu}^{k+1}J-J^{\mu}=P_{\mu}^{k+1}(J-J^{\mu}),
\qquad k\geq0.
$$
Consequently, \eqref{weightedcontraction} implies
$$
\begin{aligned}
\|F_{\lambda}^{\mu}J-J^{\mu}\|_{\xi}
&\leq(1-\lambda)\sum_{k=0}^{\infty}\lambda^k
\|P_{\mu}^{k+1}(J-J^{\mu})\|_{\xi}\\
&\leq q_{\lambda}\|J-J^{\mu}\|_{\xi}.
\end{aligned}
$$
Since there are only finitely many policies, the constant
$$
C_0=\max_{\mu}\|J^{\mu}\|_{\xi}
$$
is finite. Hence, uniformly over all policies,
$$
\|F_{\lambda}^{\mu}J\|_{\xi}
\leq q_{\lambda}\|J\|_{\xi}+(1+q_{\lambda})C_0.
$$
The conditional second-moment bound in Lemma~\ref{lmtdnoise}, together with equivalence of norms in $\mathbb{R}^n$, therefore allows us to apply Proposition 4.7 on p.~159 of \cite{BertsekasTsitsiklis96}. It follows that $\{J_t\}$ is bounded almost surely.
\end{proof}

\begin{Lemma}\label{lmct2}
For the sequence $c_t$ defined in \eqref{dfct},
$$
\limsup_{t\to\infty}c_t\leq 0,
$$
and consequently $c_t^+\to 0$.
\end{Lemma}
\begin{proof}
Let
$$
r_t=TJ_t-J_t
$$
and define its positive weighted maximum by
$$
a_t=\max\left\{0,\max_i\frac{r_t(i)}{\xi(i)}\right\}.
$$
Then $r_t\leq a_t\xi$. Recall that
$$
T_{\mu_t}J=g_{\mu_t}+P_{\mu_t}J,
\qquad \forall J.
$$
By Proposition~\ref{propdp}(d), the sequence $\{T_{\mu_t}^{k+1}J_t\}_{k\geq0}$ is bounded for each fixed $t$. Using the affine form of $T_{\mu_t}$ and Lemma~\ref{lmtmu}, we obtain
\begin{equation*}
\begin{aligned}
TJ_{t+1}
&\leq T_{\mu_t}J_{t+1}\\
&=T_{\mu_t}\bigl((1-\gamma_t)J_t
 +\gamma_tF_{\lambda}^{\mu_t}J_t+\gamma_t\omega_t\bigr)\\
&=J_{t+1}+(1-\gamma_t)(T_{\mu_t}J_t-J_t)
 +\gamma_tH_tJ_t+\gamma_tv_t,
\end{aligned}
\end{equation*}
where
$$
H_tJ_t=T_{\mu_t}(F_{\lambda}^{\mu_t}J_t)
       -F_{\lambda}^{\mu_t}J_t,
\qquad
v_t=P_{\mu_t}\omega_t-\omega_t.
$$
Since $\mu_t$ is greedy, $T_{\mu_t}J_t=TJ_t$, and hence
\begin{equation}\label{iterative}
r_{t+1}
\leq(1-\gamma_t)r_t+\gamma_tH_tJ_t+\gamma_tv_t.
\end{equation}

For a fixed policy $\mu$ and vector $J$, affine linearity gives
$$
T_{\mu}^{k+2}J-T_{\mu}^{k+1}J
=P_{\mu}^{k+1}(T_{\mu}J-J).
$$
Using this identity, \eqref{Flambda}, and $T_{\mu_t}J_t=TJ_t$, we have
$$
H_tJ_t
=(1-\lambda)\sum_{k=0}^{\infty}\lambda^kP_{\mu_t}^{k+1}r_t.
$$
By \eqref{weightedcontraction},
$$
H_tJ_t
\leq q_{\lambda}a_t\xi,
\qquad
q_{\lambda}=\frac{(1-\lambda)\beta}{1-\lambda\beta}<1.
$$
Define the monotone mapping $\Psi:\mathbb{R}^n\to\mathbb{R}^n$ by
$$
\Psi(Y)=q_{\lambda}\xi
\max\left\{0,\max_i\frac{Y(i)}{\xi(i)}\right\}.
$$
The scalar functional inside the definition of $\Psi$ is $1$-Lipschitz with respect to
$\|\cdot\|_{\xi}$. Therefore,
$$
\|\Psi(Y)-\Psi(\bar Y)\|_{\xi}
\leq q_{\lambda}\|Y-\bar Y\|_{\xi},
$$
so $\Psi$ is a contraction with unique fixed point $0$.

Lemma~\ref{lmtdnoise} gives $E[v_t\mid\mathcal{F}_t]=0$. In addition, \eqref{weightedcontraction} gives
$$
\|v_t\|_{\xi}\leq(1+\beta)\|\omega_t\|_{\xi}.
$$
Thus Lemma~\ref{lmtdnoise} and equivalence of norms imply
$$
E[\|v_t\|_{\xi}^2\mid\mathcal{F}_t]
\leq A'+B'\|J_t\|_{\xi}^2
$$
for suitable constants $A',B'>0$. Lemma~\ref{lmbounded2} implies that both $\{J_t\}$ and $\{r_t\}$ are almost surely bounded. Applying Lemma~\ref{lmcompare} with
$$
X_t=r_t,
\qquad
Q_t=J_t,
\qquad
\Phi=\Psi,
\qquad
\sigma=0,
$$
gives
$$
\limsup_{t\to\infty}r_t\leq0
$$
componentwise. Since the state space is finite and all components of $\xi$ are positive, it follows that $a_t\to0$. Consequently,
$\limsup_{t\to\infty}c_t\leq0$ and $c_t^+\to0$.
\end{proof}

Compared with \eqref{updaterule}, \eqref{updaterule2} replaces $J^{\mu_t}$ by $F_{\lambda}^{\mu_t}J_t$. Corresponding to Lemma~\ref{lmjmuest}, we establish the following result.
\begin{Lemma}\label{lmjmuest2}
Almost surely, for every $\epsilon>0$, there exists a finite time $t(\epsilon)$ such that, for all $t\geq t(\epsilon)$,
\begin{equation}\label{keyest2}
F_{\lambda}^{\mu_t}J_t\leq TJ_t+\epsilon e.
\end{equation}
\end{Lemma}
\begin{proof}
Fix $\epsilon>0$. The proof of Lemma~\ref{lmjmuest} uses only the almost-sure boundedness of $J_t$, the greediness relation $T_{\mu_t}J_t=TJ_t$, and the convergence $c_t^+\to0$. Lemmas~\ref{lmbounded2} and~\ref{lmct2} show that these properties also hold for the present sequence. Hence, almost surely, there is a finite time after which
\begin{equation}\label{td-policy-value-bound}
J^{\mu_t}\leq TJ_t+\frac{\epsilon}{2}e.
\end{equation}
For each positive integer $m$, let
$$
A_m=\left\{\sup_t\|J_t\|\leq m\right\}.
$$
On $A_m$, choose $K=K(\epsilon/2,m)$ from Lemma~\ref{lmapprox}. For every $k\geq K$,
$$
T_{\mu_t}^{k+1}J_t
\leq J^{\mu_t}+\frac{\epsilon}{2}e
\leq TJ_t+\epsilon e
$$
for all sufficiently large $t$, where the second inequality uses \eqref{td-policy-value-bound}. For $0\leq k<K$, the argument leading to \eqref{finitehorizonresidual}, which uses only greediness and Lemma~\ref{lmmon}, gives
$$
T_{\mu_t}^{k+1}J_t
\leq TJ_t+k c_t^+e.
$$
Since $c_t^+\to0$, the right-hand side is at most $TJ_t+\epsilon e$ for all sufficiently large $t$, uniformly over $0\leq k<K$. Thus, on $A_m$, eventually
$$
T_{\mu_t}^{k+1}J_t\leq TJ_t+\epsilon e,
\qquad \forall k\geq0.
$$
Averaging with the nonnegative weights $(1-\lambda)\lambda^k$, which sum to one, gives
$$
F_{\lambda}^{\mu_t}J_t\leq TJ_t+\epsilon e.
$$
Because $\bigcup_{m\geq1}A_m$ has probability one by Lemma~\ref{lmbounded2}, the conclusion follows for the fixed $\epsilon$. Taking a countable intersection over positive rational $\epsilon$ gives the stated almost-sure assertion for every $\epsilon>0$.
\end{proof}

Having established these preliminary results, we now prove the main theorem.
\begin{proof}[Proof of Theorem~\ref{main2}]
The case $\lambda=0$ follows from the weighted contraction of $T$, as noted above, and the case $\lambda=1$ is Theorem~\ref{main}. Assume $0<\lambda<1$.

For the upper bound, fix $\epsilon>0$. Lemma~\ref{lmjmuest2} gives, for all sufficiently large $t$,
$$
J_{t+1}
\leq(1-\gamma_t)J_t+\gamma_t(TJ_t+\epsilon e)+\gamma_t\omega_t.
$$
Lemma~\ref{lmbounded2}, Lemma~\ref{lmtdnoise}, and Lemma~\ref{lmcompare}, applied with
$$
X_t=J_t,
\qquad
Q_t=J_t,
\qquad
\Phi=H_{\epsilon},
\qquad
\sigma=t(\epsilon),
$$
give
$$
\limsup_{t\to\infty}J_t\leq J_{\epsilon},
$$
where $J_{\epsilon}$ is the unique fixed point of $H_{\epsilon}$. Apply this comparison for $\epsilon=1/m$, take the countable intersection of the resulting probability-one events, and then use \eqref{perturbedfixedpoint} as $m\to\infty$. We obtain
\begin{equation}\label{td-upper}
\limsup_{t\to\infty}J_t\leq J^*.
\end{equation}

It remains to establish the lower bound. For every vector $J$ and every policy $\mu$ greedy with respect to $J$,
$$
T_{\mu}J=TJ.
$$
Because $T_{\mu}x\geq Tx$ for every vector $x$, induction and monotonicity imply
\begin{equation}\label{policy-optimal-iterate}
T_{\mu}^{k+1}J\geq T^{k+1}J,
\qquad \forall k\geq0.
\end{equation}
Define
$$
G_{\lambda}J=(1-\lambda)\sum_{k=0}^{\infty}\lambda^kT^{k+1}J.
$$
This series is well defined because Proposition~\ref{propdp}(b) implies that $\{T^{k+1}J\}$ is bounded. By \eqref{policy-optimal-iterate},
$$
F_{\lambda}^{\mu_t}J_t\geq G_{\lambda}J_t.
$$
The operator $G_{\lambda}$ is monotone. Moreover, \eqref{weightedcontraction} gives
$$
\begin{aligned}
\|G_{\lambda}J-G_{\lambda}\bar J\|_{\xi}
&\leq(1-\lambda)\sum_{k=0}^{\infty}\lambda^k\beta^{k+1}
\|J-\bar J\|_{\xi}\\
&=q_{\lambda}\|J-\bar J\|_{\xi},
\end{aligned}
$$
where
$$
q_{\lambda}=\frac{(1-\lambda)\beta}{1-\lambda\beta}<1.
$$
Thus $G_{\lambda}$ is a contraction. Since $T^{k+1}J^*=J^*$ for every $k$, its unique fixed point is $J^*$.

Apply the lower-comparison part of Lemma~\ref{lmcompare} with
$$
X_t=J_t,
\qquad
Q_t=J_t,
\qquad
\Phi=G_{\lambda},
\qquad
\sigma=0.
$$
Using Lemmas~\ref{lmbounded2} and~\ref{lmtdnoise}, we obtain
\begin{equation}\label{td-lower}
\liminf_{t\to\infty}J_t\geq J^*.
\end{equation}
Combining \eqref{td-upper} and \eqref{td-lower} proves that $J_t\to J^*$ almost surely.
\end{proof}

\appendix
\section{Auxiliary stochastic-approximation estimates}

The first three lemmas establish moment bounds for the termination times and simulation noise. The final lemma gives a localization and comparison principle for stochastic approximation.

\begin{Lemma}\label{lmtau}
Let
$$
\tau_i^{\mu}=\inf\{k\geq0:X_k^{\mu}=0\}
$$
be the termination time of a trajectory starting from state $i$ under policy $\mu$. There is a finite constant $C_{\tau}$ such that
$$
\sup_{\mu}\max_{i\in S}E[(\tau_i^{\mu})^2]\leq C_{\tau}.
$$
\end{Lemma}
\begin{proof}
The block estimate preceding Assumption~\ref{assumption} and the definition of $\rho$ give
$$
P(\tau_i^{\mu}>k)\leq \rho^{\lfloor k/n\rfloor},
\qquad \forall i,\ \forall\mu,\ \forall k\geq0.
$$
For every nonnegative integer-valued random variable $\tau$,
$$
E[\tau^2]=\sum_{k=0}^{\infty}(2k+1)P(\tau>k).
$$
Therefore, uniformly over the initial state and policy,
$$
E[(\tau_i^{\mu})^2]
\leq
\sum_{k=0}^{\infty}(2k+1)\rho^{\lfloor k/n\rfloor}
=:C_{\tau}<\infty.
$$
\end{proof}

\begin{Lemma}\label{lmmcnoise}
The Monte Carlo noise in \eqref{updaterule} satisfies
$$
E[\omega_t\mid\mathcal{F}_t]=0
$$
and
$$
E[\|\omega_t\|^2\mid\mathcal{F}_t]\leq C
$$
for a constant $C$ independent of $t$.
\end{Lemma}
\begin{proof}
The fixed tie-breaking rule makes $\mu_t$ measurable with respect to $\mathcal{F}_t$. Conditional on $\mathcal{F}_t$, the trajectory used for state $i$ is therefore a trajectory under the fixed policy $\mu_t$. Lemma~\ref{lmtau} makes its cumulative cost absolutely integrable, and hence
$$
E[R_t(i)\mid\mathcal{F}_t]=J^{\mu_t}(i).
$$
Thus $E[\omega_t(i)\mid\mathcal{F}_t]=0$. If $\tau_t(i)$ is the corresponding termination time, then
$$
|R_t(i)|\leq \bar g\,\tau_t(i).
$$
Lemma~\ref{lmtau} consequently gives
$$
E[|\omega_t(i)|^2\mid\mathcal{F}_t]
=\operatorname{Var}(R_t(i)\mid\mathcal{F}_t)
\leq \bar g^2 C_{\tau}.
$$
Finally, $\|\omega_t\|^2\leq\sum_{i=1}^{n}|\omega_t(i)|^2$, so the vector bound holds with $C=n\bar g^2C_{\tau}$.
\end{proof}

\begin{Lemma}\label{lmtdnoise}
For every fixed $\lambda\in[0,1)$, the sampling noise in \eqref{updaterule2} satisfies
$$
E[\omega_t\mid\mathcal{F}_t]=0
$$
and, for constants $A,B>0$ independent of $t$,
$$
E[\|\omega_t\|^2\mid\mathcal{F}_t]
\leq A+B\|J_t\|^2.
$$
\end{Lemma}
\begin{proof}
Fix a state $i$ and define
$$
Z_{t,k}(i)=\sum_{l=0}^{k}g(i_l,\mu_t(i_l))+J_t(i_{k+1}).
$$
Conditional on $\mathcal{F}_t$, the policy $\mu_t$ and vector $J_t$ are fixed, and
$$
E[Z_{t,k}(i)\mid\mathcal{F}_t]
=(T_{\mu_t}^{k+1}J_t)(i).
$$
If $\tau_t(i)$ is the termination time of this trajectory, then, because costs after termination are zero and $J_t(0)=0$,
$$
|Z_{t,k}(i)|\leq \bar g\,\tau_t(i)+\|J_t\|,
\qquad \forall k\geq0.
$$
The right-hand side is conditionally integrable by Lemma~\ref{lmtau}. Conditional dominated convergence therefore justifies interchanging conditional expectation and the infinite weighted sum, giving
$$
\begin{aligned}
E[\widehat F_{\lambda,t}(i)\mid\mathcal{F}_t]
&=(1-\lambda)\sum_{k=0}^{\infty}\lambda^k
E[Z_{t,k}(i)\mid\mathcal{F}_t]\\
&=(F_{\lambda}^{\mu_t}J_t)(i).
\end{aligned}
$$
Thus $E[\omega_t\mid\mathcal{F}_t]=0$. Moreover,
$$
|\widehat F_{\lambda,t}(i)|
\leq \bar g\,\tau_t(i)+\|J_t\|,
$$
so
$$
\begin{aligned}
E[|\omega_t(i)|^2\mid\mathcal{F}_t]
&=\operatorname{Var}(\widehat F_{\lambda,t}(i)\mid\mathcal{F}_t)\\
&\leq E[|\widehat F_{\lambda,t}(i)|^2\mid\mathcal{F}_t]\\
&\leq 2\bar g^2C_{\tau}+2\|J_t\|^2.
\end{aligned}
$$
Since $\|\omega_t\|^2\leq\sum_{i=1}^{n}|\omega_t(i)|^2$, the stated vector bound follows, for example, with
$$
A=2n\bar g^2C_{\tau},
\qquad
B=2n.
$$
\end{proof}

\begin{Lemma}\label{lmcompare}
Let $\Phi:\mathbb{R}^n\to\mathbb{R}^n$ be monotone and a contraction in $\|\cdot\|_{\xi}$, with unique fixed point $x^*$. Let $\{X_t\}$ and $\{Q_t\}$ be adapted and almost surely bounded, and let $\{\eta_t\}$ satisfy
$$
E[\eta_t\mid\mathcal{F}_t]=0,
\qquad
E[\|\eta_t\|_{\xi}^2\mid\mathcal{F}_t]
\leq A+B\|Q_t\|_{\xi}^2.
$$
Suppose that there is an almost surely finite integer-valued random time $\sigma$ such that
\begin{equation}\label{localized-upper}
X_{t+1}\leq(1-\gamma_t)X_t+\gamma_t\Phi(X_t)+\gamma_t\eta_t,
\qquad \forall t\geq\sigma.
\end{equation}
Then
$$
\limsup_{t\to\infty}X_t\leq x^*
$$
componentwise. If the inequality in \eqref{localized-upper} is reversed, then
$$
\liminf_{t\to\infty}X_t\geq x^*.
$$
\end{Lemma}
\begin{proof}
For positive integers $m$ and $s$, set
$$
A_{m,s}=
\left\{
\sigma\leq s,\ 
\sup_t\|X_t\|_{\xi}\leq m,\ 
\sup_t\|Q_t\|_{\xi}\leq m
\right\}.
$$
The union of these events over $m$ and $s$ has probability one. Define the stopping time
$$
\tau_m=\inf\{t:\|Q_t\|_{\xi}>m\},
$$
with the convention that the infimum of the empty set is $\infty$, and define the stopped noise
$$
\eta_t^{(m)}=\eta_t\mathbf{1}_{\{t<\tau_m\}}.
$$
Then $\{\eta_t^{(m)}\}$ remains a martingale-difference sequence and
$$
E[\|\eta_t^{(m)}\|_{\xi}^2\mid\mathcal{F}_t]\leq A+Bm^2.
$$
Let $\Pi_m$ denote componentwise projection onto the weighted box
$\{x:\|x\|_{\xi}\leq m\}$. For fixed $m$ and $s$, define
$$
Y_s^{m,s}=\Pi_m(X_s)
$$
and, for $t\geq s$,
$$
Y_{t+1}^{m,s}
=(1-\gamma_t)Y_t^{m,s}
+\gamma_t\Phi(Y_t^{m,s})
+\gamma_t\eta_t^{(m)}.
$$
Proposition 4.4 on p.~156 of \cite{BertsekasTsitsiklis96} gives
$Y_t^{m,s}\to x^*$ almost surely. On $A_{m,s}$, we have
$\tau_m=\infty$, $Y_s^{m,s}=X_s$, and \eqref{localized-upper} holds for every $t\geq s$. Monotonicity of $\Phi$ and $0<\gamma_t\leq1$ therefore imply inductively that
$$
X_t\leq Y_t^{m,s},\qquad \forall t\geq s.
$$
Hence $\limsup_tX_t\leq x^*$ on $A_{m,s}$. Taking the countable union over $m$ and $s$ proves the upper comparison. The lower comparison is identical with all inequalities reversed.
\end{proof}

\bibliographystyle{amsplain}
\bibliography{ssp}

\end{document}